\documentclass{article}

\usepackage{spconf,amsmath,graphicx}
\usepackage{graphics}
\usepackage[named]{algo}
\usepackage{stfloats}
\usepackage{amssymb}
\usepackage{amsmath}
\usepackage{amsfonts}
\usepackage{subfigure}
\usepackage{float}
\usepackage{pifont}
\usepackage{setspace}
\usepackage{url}
\usepackage{paralist}
\usepackage[perpage,symbol*]{footmisc}

\newcommand{\Section}{\section}
\newcommand{\SubSection}{\subsection}

\renewcommand{\thefootnote}{}

\title{Group-sparse Matrix Recovery}
\name{Xiangrong Zeng\ \  and \  M\'{a}rio A. T. Figueiredo
}
\address{Instituto de Telecomunica\c{c}\~oes, Instituto Superior T\'ecnico, Lisboa, Portugal}
\begin{document}
%
\maketitle
\let\thefootnote\relax\footnotetext{\hspace{-0.75cm} Work partially supported by Funda\c{c}\~{a}o para a Ci\^{e}ncia  e Tecnologia, grants PEst-OE/EEI/LA0008/2013
and PTDC/EEI-PRO/1470/2012.}

\begin{abstract}
We apply the OSCAR (octagonal selection and clustering algorithms for regression)
in recovering group-sparse matrices (two-dimensional---2D---arrays) from compressive
measurements.
We propose a 2D version of OSCAR (2OSCAR) consisting of the $\ell_1$
norm and the pair-wise $\ell_{\infty}$ norm, which is convex but non-differentiable.
We show that the proximity operator of 2OSCAR can be computed based on that of
OSCAR. The 2OSCAR problem can thus be efficiently solved by state-of-the-art
proximal splitting algorithms. Experiments on group-sparse
2D array recovery show that 2OSCAR regularization solved by the SpaRSA algorithm
is the fastest choice, while the PADMM algorithm (with debiasing) yields the most
accurate results.
\end{abstract}
\Section{Introduction}\label{sec:intro}

The problem studied in this paper is the classical one of recovering ${\bf X}$ from
\begin{equation}\label{linearmodel}
 {\bf Y}={\bf A}{\bf X} + {\bf W},
\end{equation}
where ${\bf A}\in\mathbb{R}^{m\times n}$ is a known sensing matrix,
${\bf X}\in\mathbb{R}^{n\times d}$ the original unknown matrix/2D-array,
${\bf Y}\in\mathbb{R}^{m\times d}$ is the observed data, and ${\bf W}\in\mathbb{R}^{m\times d}$
denotes additive noise. In many cases of interest, we have $ m < n $, making \eqref{linearmodel} an ill-posed problem,
which can only be addressed by using some
form of regularization that injects prior knowledge about the unknown $\bf X$.
Classical regularization formulations seek solutions of problems
of the form
\begin{equation}\label{regularationmodel1}
\min_{\bf X}  {\it F}(\bf X) + \Phi(\bf X),
\end{equation}
or one of the equivalent (under mild conditions) forms
\begin{equation}\label{regularationmodel2}
\min_{\bf X} \Phi({\bf X}) \:\mbox{ s.t.} \: {\it F}({\bf X}) \leq \varepsilon \  {\mbox{or}} \
\min_{\bf X} F({\bf X}) \:\mbox{ s.t.} \: \Phi({\bf X}) \leq \epsilon,
\end{equation}
where $F({\bf X}) $ is the data-fidelity term and $\Phi(\bf X)$ is the regularizer,
the purpose of which is to enforce certain properties on ${\bf X}$, such as sparsity
or group sparsity, and $\varepsilon$ and $\epsilon$ are positive parameters.

Problem (\ref{linearmodel}) is more challenging than the more studied
linear inverse problem of recovering a vector ${\bf x}\in\mathbb{R}^{n}$ from
\begin{equation}\label{linearmodel2}
 {\bf y}={\bf A}{\bf x} +{\bf w},
\end{equation}
where ${\bf A}\in\mathbb{R}^{m\times n}$ is also a known sensing matrix,
${\bf y}\in\mathbb{R}^{m}$ is the observed vector, and ${\bf w}\in\mathbb{R}^{m\times d}$ is
additive noise. Comparing with (\ref{linearmodel2}), the matrix ${\bf X}$ of interest
in (\ref{linearmodel}) is always assumed to be, not only sparse, but also to have a
particular sparse structure. For instance, in the {\it multiple measurement vector} model \cite{cotter2005sparse},
\cite{rao1998sparse},  \cite{zhang2011sparse}, ${\bf X}$ is an unknown source matrix that
 should be row sparse; in group LASSO \cite{yuan2005model}, \cite{qin2012structured},
\cite{liu2010fast}, ${\bf X}$ is a coefficient matrix that is also enforced to be row sparse;
in {\it multi-task learning} \cite{zhou2011clustered}, \cite{zhou2011multi}, ${\bf X}$ is a
task parameter matrix, which is usually assumed to be row or/and column sparse.
In this paper, we pursue more general sparsity patterns for ${\bf X}$, that is,
the arrangement of each group of nonzeros in ${\bf X}$ is not limited to rows and/or columns,
but may include row/columns segments, blocks, or other groups of connected non-zero elements.
Before addressing the question of whether or not there are any available regularizers able to
promote this kind of group sparsity, we first briefly review existing group-sparsity-inducing
regularizers.

In recent years, much attention has been devoted not only to the sparsity of solutions,
but also the structure of this sparsity \cite{Bach2012}. In other words, not only the
number of non-zeros in the solutions, but also how these non-zeros are located, are of interest.
This research direction has lead to the concept of group/block sparsity \cite{yuan2005model},
\cite{eldar2009block}, or more general structured sparsity patterns \cite{huang2011learning},
\cite{micchelli2010regularizers}, \cite{mairal2010network}.
A classical model for group sparsity is the {\it group LASSO} \cite{yuan2005model}, which, making
use of more information than the original LASSO \cite{tibshirani1996regression} (namely, the
structure of the groups) is able to simultaneously encourage sparsity and group sparsity.
In addition, the {\it sparse group LASSO} approach was proposed in \cite{simon2012sparse}; its regularizer
consists of an $\ell_1$ term plus the group LASSO regularizer, thus unlike group LASSO,
it not only selects groups, but also individual variables within each group.

It has also been observed that in some real-world problems, it makes sense to encourage the
solution, not only to be sparse, but also to have several components sharing similar values.
To formalize this goal, several generic models have been proposed, such as the {\it elastic net}
\cite{zou2005regularization}, the {\it fused LASSO} \cite{tibshirani2004sparsity}, and the
{\it octagonal shrinkage and clustering algorithm for regression} (OSCAR) \cite{bondell2007simultaneous}.

\begin{figure}
	\centering
		\includegraphics[width=0.7\columnwidth]{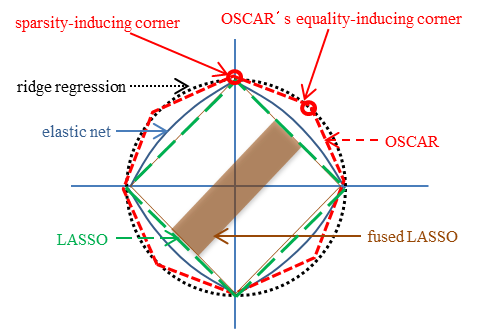}
	\caption{Illustration of LASSO, elastic net, fused LASSO and OSCAR}
	\label{fig:figure1}
\end{figure}

The level curves of several of the regularizers mentioned in the previous paragraph
(for the 2D case) are shown in Fig.~\ref{fig:figure1}. The figure illustrates
why these models promote variable grouping (unlike LASSO). Firstly, the regularizer
of the elastic net \cite{zou2005regularization} consists of a $\ell_1$ term and
a $\ell_2$ term, thus simultaneously promoting sparsity and group-sparsity,
in which the former comes from the sparsity-inducing corners (see Fig. 1) while the latter
from its strictly convex edges, which creates a grouping effect similar to a quadratic/ridge
regularizer. Secondly, the regularizer of the fused LASSO is composed of a $\ell_1$ term and
a total variation term, which encourages successive variables (in a certain order)
to be similar, making it able to promote both sparsity and smoothness.
Thirdly, the OSCAR regularizer (proposed by Bondell and Reich \cite{bondell2007simultaneous})
is constituted by a $\ell_1$ term and a pair-wise $\ell_{\infty}$ term, which promotes
equality (in absolute value) of each pair of variables.

There are some recent variants of the above group-sparsity regularizers,
such as the {\it weighted fused LASSO}, presented in \cite{daye2009shrinkage}.
The {\it pair-wise fused LASSO} \cite{daye2009shrinkage}, which
uses the pair-wise term of OSCAR, extends the fused LASSO to cases where
the variables have no natural ordering. A novel {\it graph-guided fused
LASSO} was proposed in \cite{kim2009multivariate}, where the grouping
structure is modeled by a graph. A Bayesian version of the elastic net was
developed in \cite{li2010bayesian}. Finally, an adaptive grouping
pursuit method was proposed in \cite{shen2010grouping}, but the underlying
regularizer is neither sparsity-promoting nor convex.

The fused LASSO, elastic net, and OSCAR regularizers all have the
ability  to promote sparsity and variable grouping. However, as
pointed out in \cite{zhong2012efficient}, OSCAR outperforms the other
two models in terms of grouping. Moreover, the fused LASSO is not
suitable for group according to magnitude, and the grouping ability
of the convex edges of the elastic net is inferior to that of OSCAR.
Thus, this paper will focus on the OSCAR regularizer to solve
the problems of group-sparse matrix recovery.

In this paper, we will propose a two-dimensional (matrix) version of OSCAR (2OSCAR)
for group-sparse matrix recovery. Solving OSCAR regularization problems has
been addressed in our previous work \cite{zeng2013solving}, in which,
six state-of-the-art proximal splitting algorithms: FISTA \cite{beck2009fast},
TwIST \cite{bioucas2007new}, SpaRSA \cite{wright2009sparse}, ADMM \cite{boyd2011distributed},
 SBM \cite{goldstein2009split} and PADMM \cite{chambolle2011first} are investigated.
Naturely, we build the relationship between OSCAR and 2OSCAR, and then address
2OSCAR regularization problems as in \cite{zeng2013solving}.

\SubSection*{Terminology and Notation}

We denote vectors or general variables by lower case letters, and matrices by upper case ones.
The $\ell_1$ norm of a vector ${\bf x}\in\mathbb{R}^{n}$ is $\|{\bf x}\|_1=
\sum_{i=1}^n\left|{\bf x}_i\right|$ where ${\bf x}_i$ represents the $i$-th component
of ${\bf x}$, and that of a matrix ${\bf X}\in\mathbb{R}^{n\times d}$ is $\|{\bf X}\|_1=
\sum_{i=1}^n\sum_{j=1}^d\left|{\bf X}_{(i,j)}\right|$   where ${\bf X}_{(i,j)}$
the entry of ${\bf X}$ at the $i$-th row and the $j$-th column. Let $\|{\bf X}\|_F=
\bigl(\sum_{i=1}^n\sum_{j=1}^d{\bf X}_{(i,j)}^2 \bigr)^{1/2}$ be the Frobenius norm
of ${\bf X}$.

We now briefly review some elements of convex analysis that will be used below.
Let $\mathcal{H}$ be a real Hilbert space with inner product $\langle\cdot,\cdot\rangle$
and norm $\left\|\cdot\right\|$. Let $f:\mathcal{H}\rightarrow \left[-\infty,+\infty\right]$
be a function and $\Gamma$ be the class of all lower semi-continuous, convex, proper functions
(not equal to $+\infty$ everywhere and never equal to $-\infty$). The proximity operator
\cite{bauschke2011convex} of $\lambda\, f$ (where $f\in \Gamma$ and $\lambda \in \mathbb{R}_+$) is defined as
\begin{equation}\label{proximityoperator}
\mbox{prox}_{\lambda f} \left( {\bf v}\right) =
\arg\min_{{\bf x}\in \mathcal{H}} \left( \lambda f\left( {\bf x} \right)
+ \frac{1}{2} \left\|{\bf x}-{\bf v}\right\|^2\right).
\end{equation}

\Section{Recovering Group-sparse Matrices}\label{sec:OSCARmOSCAR}

\SubSection{OSCAR and its 2D Version (2OSCAR)} \label{subsec:OSCAR}

\begin{figure} [htbp]
	\centering
		\includegraphics[width=0.7\columnwidth]{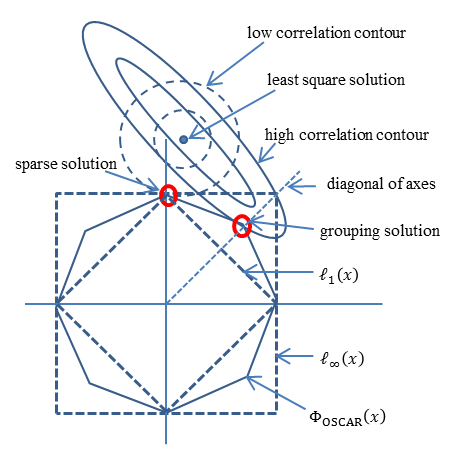}
	\caption{Illustration of the $\ell_2$ term and  $\Phi_{\mbox{\tiny OSCAR}} \left( {\bf x}\right)$
	in the $n = 2$ case. 	In this example, for the same least square solution $({\bf A}^T{\bf A})^{-1}{\bf A}^T{\bf y}$,
		the high correlation contour is more likely to hit the equality-inducing (grouping) vertex,
		whereas the low correlation contour ``prefers" the sparsity-inducing vertex. }
	\label{fig:OSCAR}
\end{figure}

The OSCAR criterion is given by \cite{bondell2007simultaneous}
\begin{equation}\label{OSCAR}
\min_{{\bf x}\in\mathbb{R}^{ n}} \frac{1}{2} \left\| {\bf y} - {\bf A}{\bf x} \right\|_2^2 +
\underbrace{\lambda_1 \left\|{\bf x} \right\|_1 + \lambda_2 \sum_{i<j}
\max \left\{ |x_i |,  |x_j |\right\}}_{\Phi_{\mbox{\tiny OSCAR}} \left( {\bf x}\right)},
\end{equation}
where the $\ell_2$ term seeks data-fidelity, while the regularizer
$\Phi_{\mbox{\tiny OSCAR}} \left( {\bf x}\right)$ consists of an $\ell_1$ term
(promoting sparsity) and a pair-wise $\ell_\infty$ term ($(n(n-1))/2$ pairs in total)
encouraging  equality (in magnitude) of each pair of elements $|x_i|$ and $|x_j|$.
Thus, $\Phi_{\mbox{\tiny OSCAR}} \left( {\bf x}\right)$ promotes both sparsity and grouping.
Parameters $\lambda_1$ and $\lambda_2$ are nonnegative constants controlling
the relative weights of the two terms. If $\lambda_2=0$, \eqref{OSCAR} becomes the
LASSO, while if $\lambda_1=0$,  $\Phi_{\mbox{\tiny OSCAR}}$ becomes a pair-wise
$\ell_\infty$ regularizer. Note that, for any choice of $\lambda_1, \lambda_2\in\mathbb{R}_+$,
$\Phi_{\mbox{\tiny OSCAR}} \left( {\bf x}\right)$ is convex and its ball is octagonal in the 2-D case.
In the 2D case, the 8 vertices of this octagon can be divided into two categories: four sparsity-inducing
vertices (located on the axes) and four equality-inducing vertices (see Fig.~\ref{fig:figure1}).
Fig. \ref{fig:OSCAR} depicts the a data-fidelity term and $\Phi_{\mbox{\tiny OSCAR}} \left( {\bf x}\right)$,
illustrating its possible effects.

As discussed in Section \ref{sec:intro}, compared with group LASSO,
OSCAR doesn't require a pre-specification of group structure; compared with
fused LASSO, it doesn't depend on a certain order of the variables;
compared with the elastic net, it has a stronger equality-inducing
ability. All these features make OSCAR a convenient regularizer in
many applications. A fundamental building block for using OSCAR
is its proximity operator, $\mbox{prox}_{\Phi_{\mbox{\tiny OSCAR}}}$,
which can be obtained exactly or approximately by the
{\it grouping proximity operator} and the
{\it approximate proximity operator} proposed in \cite{zeng2013solving},
respectively.

To address the matrix inverse problem (\ref{linearmodel}), we
propose a matrix version of OSCAR, termed 2OSCAR, given by
\begin{equation}\label{mOSCAR}
\min_{{\bf X}\in\mathbb{R}^{n\times d}} \frac{1}{2} \left\| {\bf Y} - {\bf A}{\bf X} \right\|_F^2
+ \Phi_{\mbox{\tiny 2OSCAR}} \left( {\bf X}\right)
\end{equation}
where ${\bf A}\in\mathbb{R}^{m\times n}$, ${\bf Y}\in\mathbb{R}^{m \times d}$ and
\begin{equation}\label{mOSCARregularizerOSCAR}
\Phi_{\mbox{\tiny 2OSCAR}} \left( {\bf X}\right) = \Phi_{\mbox{\tiny OSCAR}}
\bigl( \text{vec}\left( {\bf X}\right)\bigr)
\end{equation}
with $\text{vec}$ denoting the vectorization function, which transforms a
matrix into a vector by stacking the columns on top of each other.

Observing that, for any matrix ${\bf Z}$, we have $\left\|  {\bf Z}\right\|_F^2 = \|\text{vec}({\bf Z})\|_2^2$, we
can write proximity operator of 2OSCAR as
\begin{equation}\label{POofmOSCAR}
\mbox{prox}_{\Phi_{\mbox{\tiny 2OSCAR}}} \left( {\bf {\bf Z}}\right) =
\text{vec}^{-1}\left( \mbox{prox}_{\Phi_{\mbox{\tiny OSCAR}}} \bigl( \text{vec} \left({\bf Z}\right)\bigr) \right).
\end{equation}
where $\mbox{prox}_{\Phi_{\mbox{\tiny OSCAR}}}$ can be obtained by
the algorithm proposed in \cite{zeng2013solving} and $\text{vec}^{-1}$ is the
inverse of the vectorization function, that is, it takes a $nd$ vector and
yields an $n\times d$ matrix.

\SubSection{Algorithms}\label{sec:OSCARandPSAS}

The 2OSCAR problem that needs to be solved is given by Equation~\eqref{mOSCAR}.
It is clear that the objective function is convex (since both terms are convex)
and coercive (that is, it goes to $\infty$ as $\|{\bf X}\|\rightarrow \infty$), thus
the set of minimizers is not empty.


To solve \eqref{mOSCAR}, we investigate six state-of-the-art proximal
splitting algorithms: FISTA \cite{beck2009fast}, TwIST \cite{bioucas2007new},
SpaRSA \cite{wright2009sparse}, ADMM \cite{boyd2011distributed},  SBM
\cite{goldstein2009split} and PADMM \cite{chambolle2011first}. Due to limitation of space,
we next only detail SpaRSA, since it has been experimentally shown to be the fastest one.
However, we will report below experimental results with the aforementioned six algorithms.

SpaRSA \cite{wright2009sparse} is a fast proximal splitting that uses
the step-length selection method of Barzilai and Borwein \cite{barzilai1988two}.
Its application to solve 2OSCAR problems leads to the following algorithm:

\vspace{0.05cm}
\begin{algorithm}{SpaRSA for 2OSCAR}{
\label{alg:sparsa}}
Set $k=1$, $\eta > 1$, $\alpha_0 = \alpha_{\tiny \mbox{min}} > 0$,
$\alpha_{\tiny \mbox{max}} > \alpha_{\tiny \mbox{min}}$,  and  ${\bf X}_0$.\\
     ${\bf V}_0 = {\bf X}_0 -{\bf A}^T \left({\bf A} {\bf X}_0 - Y\right)/\alpha_0$\\
     ${\bf X}_1  = \mbox{Prox}_{\Phi_{\mbox{\tiny 2OSCAR}}/\alpha_0} \left({\bf V}_0 \right)$\\
\qrepeat\\
     ${\bf S}_k = {\bf X}_k - {\bf X}_{k-1}$ \\
		 ${\bf R}_k = {\bf A}^T{\bf A}{\bf S}_k$\\
		 $\hat{\alpha}_k = \frac{\left({\bf S}_k\right)^T{\bf R}_k}{\left({\bf S}_k\right)^T {\bf S}_k}$\\
		 $\alpha_k = \max\left\{\alpha_{min}, \min\left\{\hat{\alpha}_k,\alpha_{max}\right\}\right\}$\\
     \qrepeat\\
          ${\bf V}_k = {\bf X}_k -{\bf A}^T \left({\bf A} {\bf X}_k - Y\right)/\alpha_k$\\
          ${\bf X}_{k+1}  = \mbox{Prox}_{\Phi_{\mbox{\tiny 2OSCAR}}/\alpha_k} \left({\bf V}_k \right)$\\
		      $\alpha_k \leftarrow \eta \alpha_k$
		 \quntil ${\bf X}_{k+1}$ satisfies an acceptance criterion.\\
     $k \leftarrow k + 1$
\quntil some stopping criterion is satisfied.
\end{algorithm}
\vspace{0.05cm}

A common acceptance criterion in line 13 requires the objective function to
decrease; see \cite{wright2009sparse} for details.

\SubSection{Debiasing}
As is well known, the solutions obtained under 2OSCAR (and many other types of regularizers)
are attenuated/biased in magnitude. Thus, it is common practice to apply {\it debiasing}
as a postprocessing step; {\it i.e.}, the solutions obtained by, say the the SpaRSA algorithm
provides the structure/support of the estimate and the debiasing step recovers the
magnitudes of the solutions. The debiasing method used in SpaRSA \cite{wright2009sparse}
is also adopted in this paper. Specifically, the debiasing phase solves
\begin{equation}\label{debias}
\begin{split}
\widehat{\bf X}_{\mbox{\tiny debias}} = & \arg\min_{\bf X}
\left( \left\|{\bf A}{\bf X} - {\bf Y}\right\|^2_F\right)\\
& \text{s.t.} \quad \text{supp}({\bf X}) = \text{supp}(\tilde{\bf X})
\end{split}
\end{equation}
where ${\bf \tilde{X}}$ is the estimate produced by the SpaRSA
(or any other) algorithm and $\text{supp}({\bf X})$ denotes the set of
indices of the non-zero elements of ${\bf X}$.
This problem is solved by conjugate gradient procedure; see
\cite{wright2009sparse} for more details.

\Section{Experiments} \label{sec:experiments}
All the experiments were performed using MATLAB on a 64-bit
Windows 7 PC with an Intel Core i7 3.07 GHz processor and 6.0GB
of RAM. The performance of the different algorithms is assessed
via the following five metrics, where ${\bf E}$ is an estimate of ${\bf X}$):
\begin{itemize}

	\item Mean absolute error, $\text{MAE} = \left\|{\bf X} - {\bf E}\right\|_1/(nd)$;
	\item Mean square error, $\text{MSE} = \left\|{\bf X} - {\bf E}\right\|_F^2/(nd)$;
    \item Position error rate,
	\[
	\text{PER}=\sum_{i=1}^n\sum_{j=1}^d \left|\left|\mbox{sign}\left({\bf X}_{(i,j)}
	\right)  \right| - \left|\mbox{sign}\left({\bf E}_{(i,j)} \right)  \right|  \right| /(nd).
	\]
   \item Elapsed time ({TIME}).
\end{itemize}

\begin{figure}
		\includegraphics[width=1.00\columnwidth]{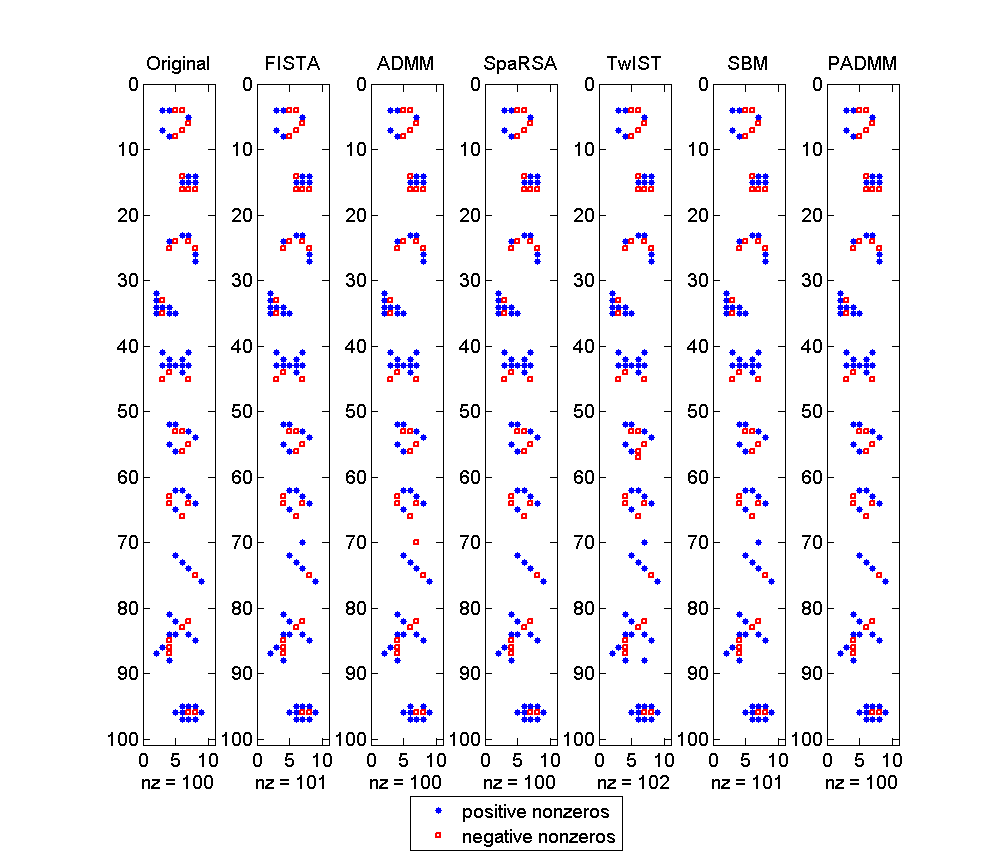}
	\caption{Original and recovered matrices}
	\label{fig:recovered_matrices}
\end{figure}

We consider the experiments on recovery of a $100 \times 10$ matrix ${\bf X}$ with different styles of groups
-- blocks, lines and curved groups, consisting of positive and negative elements.
The observed matrix ${\bf Y}$ is generated by (\ref{linearmodel}), in which the variance of the noise
${\bf W}$ is $\sigma^2=0.16$. The sensing matrix ${\bf A}$ is a $65 \times 100$ matrix with components
sampled from the standard normal distribution.  There are 100 nonzeros in the original $100 \times 10$
matrix, with values arbitrarily chosen from the set $\{-7, -8, -9, 7, 8, 9\}$ (Fig. \ref{fig:recovered_matrices}).

\begin{table} [h!]
\centering \caption{Results of metrics} \label{tab:resultsofsparserecovery}
\begin{tabular}{|l|l l|l l|l l|c|}
\hline\rule[-0.1cm]{0cm}{0.4cm}
\footnotesize Metrics & \multicolumn{2}{|c|}{\footnotesize TIME (sec.)} & \multicolumn{2}{|c|}{\footnotesize MAE} &  \multicolumn{2}{|c|}{\footnotesize MSE} &  {\footnotesize PER}\\
\hline
\footnotesize debiasing & \footnotesize yes & \footnotesize no &  \footnotesize yes &\footnotesize  no & \footnotesize yes & \footnotesize no &-\\
\hline
\footnotesize FISTA   & \footnotesize 4.37 & \footnotesize 4.26 & \footnotesize 0.0784 & \footnotesize 0.477 & \footnotesize 2.45 &\footnotesize 0.202 & \footnotesize 0.1\%  \\
\footnotesize TwIST   & \footnotesize 5.10& \footnotesize 4.45  &  \footnotesize 0.0799& \footnotesize 0.480  & \footnotesize 2.47 &\footnotesize 0.202 & \footnotesize 0.2\%  \\
\footnotesize SpaRSA   & \footnotesize 2.25& \footnotesize 2.26  &  \footnotesize 0.0784& \footnotesize 0.477  & \footnotesize 2.44 &\footnotesize 0.202 & \footnotesize 0.0\% \\
 \footnotesize ADMM & \footnotesize 6.65& \footnotesize 6.60  &  \footnotesize 0.0786& \footnotesize 0.477  & \footnotesize 2.44 &\footnotesize 0.206 & \footnotesize 0.2\%  \\
\footnotesize SBM & \footnotesize 6.32 & \footnotesize 6.22  &  \footnotesize 0.0784&   \footnotesize 0.477  & \footnotesize 2.45 &\footnotesize 0.202 & \footnotesize 0.1\%  \\
\footnotesize PADMM & \footnotesize 6.01 & \footnotesize 5.97 &  \footnotesize 0.0762& \footnotesize 0.456  & \footnotesize 2.42 &\footnotesize 0.182 & \footnotesize 0.0\%  \\
\hline
\end{tabular}
\end{table}

We run algorithms mentioned above (FISTA, TwIST, SpaRSA, SBM, ADMM, PADMM), with and without debiasing.
The stopping condition is $\left\|{\bf X}_{k+1}- {\bf X}_k\right\|/\left\|{\bf X}_{k+1}\right\|\leq 0.001$,
where ${\bf X}_k$ represents the estimate at the$k$-th iteration. We set $\lambda_1=0.5$ and $\lambda_2=0.0024$.
Other parameters are hand-tuned in each case for the best improvement in MAE.
The recovered matrices are shown in Fig. \ref{fig:recovered_matrices} and the quantitative results are 
reported in Table \ref{tab:resultsofsparserecovery}.

We can conclude from Fig. \ref{fig:recovered_matrices} and Table \ref{tab:resultsofsparserecovery}
that the 2OSCAR criterion solved by proximal splitting algorithms with debiasing is able to accurately
recover group-sparse matrices.  Among the algorithms, the SpaRSA is the fastest,
while the PADMM obtains the most accurate solutions.

\Section{Conclusions}
\label{sec:conclusions}

We have applied the OSCAR regularizer to recover group-sparse matrix with arbitrary groups
from compressive measurements. A matrix version of the OSCAR (2OSCAR) has been proposed 
and solved by six state-of-the-art proximal spiting algorithms:
FISTA, TwIST, SpaRSA, SBM, ADMM and PADMM, with or without debaising. Experiments on 
group-sparse matrix recovery show that the 2OSCAR regularizer solved 
by the SpaRSA algorithm has the fastest convergence,  while the PADMM leads to 
the most accurate estimates.


\begin{thebibliography}{10}

\bibitem{cotter2005sparse}
S.F. Cotter, B.D. Rao, K.~Engan, and K.~Kreutz-Delgado,
\newblock ``Sparse solutions to linear inverse problems with multiple
  measurement vectors,''
\newblock {\em IEEE Trans. on Signal Processing}, vol. 53, pp.
  2477--2488, 2005.

\bibitem{rao1998sparse}
BD~Rao and K.~Kreutz-Delgado,
\newblock ``Sparse solutions to linear inverse problems with multiple
  measurement vectors,''
\newblock in {\em Proc. of the 8th IEEE Digital Signal Processing
  Workshop}, 1998.

\bibitem{zhang2011sparse}
Z.~Zhang and B.D. Rao,
\newblock ``Sparse signal recovery with temporally correlated source vectors
  using sparse bayesian learning,''
\newblock {\em IEEE Jour. of Selected Topics in Signal Processing}, vol. 5,
  pp. 912--926, 2011.

\bibitem{yuan2005model}
M.~Yuan and Y.~Lin,
\newblock ``Model selection and estimation in regression with grouped
  variables,''
\newblock {\em Jour. of the Royal Statistical Society (B)}, vol. 68, pp.
  49--67, 2005.

\bibitem{qin2012structured}
Z.~Qin and D.~Goldfarb,
\newblock ``Structured sparsity via alternating direction methods,''
\newblock {\em The Jour. of Machine Learning Research}, vol. 98888, pp.
  1435--1468, 2012.

\bibitem{liu2010fast}
L. ~Yuan, J.~Liu and J.~Ye,
\newblock ``Efficient methods for overlapping group lasso,''
\newblock {\em IEEE Transactions on Pattern Analysis and Machine Intelligence}, vol. ~35, pp.
  ~2104--2116, 2013.

\bibitem{zhou2011clustered}
J.~Zhou, J.~Chen, and J.~Ye,
\newblock ``Clustered multi-task learning via alternating structure
  optimization,''
\newblock {\em Advances in Neural Information Processing Systems}, vol. 25,
  2011.

\bibitem{zhou2011multi}
J.~Zhou, L.~Yuan, J.~Liu, and J.~Ye,
\newblock ``A multi-task learning formulation for predicting disease
  progression,''
\newblock in {\em 17th ACM SIGKDD international Conf.
  on Knowledge discovery and Data Mining}, 2011, pp. 814--822.

\bibitem{Bach2012}
F.~Bach, R.~Jenatton, J.~Mairal, and G.~Obozinski,
\newblock ``Structured sparsity through convex optimization,''
\newblock {\em Statistical Science}, vol. 27, pp. 450--468, 2012.

\bibitem{eldar2009block}
Y.C. Eldar and H.~Bolcskei,
\newblock ``Block-sparsity: Coherence and efficient recovery,''
\newblock in {\em IEEE International Conf.  on Acoustics, Speech and Signal
  Processing (ICASSP)}, 2009, pp. 2885--2888.

\bibitem{huang2011learning}
J.~Huang, T.~Zhang, and D.~Metaxas,
\newblock ``Learning with structured sparsity,''
\newblock {\em The Jour. of Machine Learning Research}, vol.~12, 
pp.~3371--3412, 2011.

\bibitem{micchelli2010regularizers}
C.A. Micchelli, J.M. Morales, and M.~Pontil,
\newblock ``Regularizers for structured sparsity,''
\newblock {\em Advances in Computational Math.}, pp. 1--35, 2010.

\bibitem{mairal2010network}
J.~Mairal, R.~Jenatton, G.~Obozinski, and F.~Bach,
\newblock ``Convex and network flow algorithms for structured sparsity,''
\newblock {\em The Jour. of Machine Learning Research}, vol.~12, 
pp.~2681--2720, 2011.

\bibitem{tibshirani1996regression}
R.~Tibshirani,
\newblock ``Regression shrinkage and selection via the lasso,''
\newblock {\em Jour. of the Royal Statistical Society (B)}, pp. 267--288,
  1996.

\bibitem{simon2012sparse}
N.~Simon, J.~Friedman, T.~Hastie, and R.~Tibshirani,
\newblock ``The sparse-group lasso,''
\newblock {\em Jour. of Comput. and Graphical Statistics}, 2012,
\newblock to appear.

\bibitem{zou2005regularization}
H.~Zou and T.~Hastie,
\newblock ``Regularization and variable selection via the elastic net,''
\newblock {\em Jour. of the Royal Statistical Society (B)}, vol. 67, pp.
  301--320, 2005.

\bibitem{tibshirani2004sparsity}
R.~Tibshirani, M.~Saunders, S.~Rosset, J.~Zhu, and K.~Knight,
\newblock ``Sparsity and smoothness via the fused lasso,''
\newblock {\em Jour. of the Royal Statistical Society (B)}, vol. 67, pp.
  91--108, 2004.

\bibitem{bondell2007simultaneous}
H.D. Bondell and B.J. Reich,
\newblock ``Simultaneous regression shrinkage, variable selection, and
  supervised clustering of predictors with {OSCAR},''
\newblock {\em Biometrics}, vol. 64, pp. 115--123, 2007.

\bibitem{daye2009shrinkage}
Z.J. Daye and X.J. Jeng,
\newblock ``Shrinkage and model selection with correlated variables via
  weighted fusion,''
\newblock {\em Computational Statistics \& Data Analysis}, vol. 53, pp.
  1284--1298, 2009.

\bibitem{kim2009multivariate}
S.~Kim, K.A. Sohn, and E.P. Xing,
\newblock ``A multivariate regression approach to association analysis of a
  quantitative trait network,''
\newblock {\em Bioinformatics}, vol. 25, pp. i204--i212, 2009.

\bibitem{li2010bayesian}
Q.~Li and N.~Lin,
\newblock ``The bayesian elastic net,''
\newblock {\em Bayesian Analysis}, vol. 5, pp. 151--170, 2010.

\bibitem{shen2010grouping}
X.~Shen and H.C. Huang,
\newblock ``Grouping pursuit through a regularization solution surface,''
\newblock {\em Jour. of the American Statistical Assoc.}, vol. 105, pp.
  727--739, 2010.

\bibitem{zhong2012efficient}
L.W. Zhong and J.T. Kwok,
\newblock ``Efficient sparse modeling with automatic feature grouping,''
\newblock {\em IEEE Trans. on Neural Networks and Learning Systems}, vol.
  23, pp. 1436--1447, 2012.

\bibitem{zeng2013solving}
X.~Zeng and M.A.T. Figueiredo,
\newblock ``Solving {OSCAR} regularization problems by proximal splitting
  algorithms,''
\newblock {\em arXiv preprint arxiv.org/abs/1309.6301}, 2013.

\bibitem{beck2009fast}
A.~Beck and M.~Teboulle,
\newblock ``A fast iterative shrinkage-thresholding algorithm for linear
  inverse problems,''
\newblock {\em SIAM Jour. on Imaging Sciences}, vol. 2, pp. 183--202, 2009.

\bibitem{bioucas2007new}
J.M. Bioucas-Dias and M.A.T. Figueiredo,
\newblock ``A new twist: two-step iterative shrinkage/thresholding algorithms
  for image restoration,''
\newblock {\em IEEE Trans. on Image Processing}, vol. 16, pp. 2992--3004,
  2007.

\bibitem{wright2009sparse}
S.J. Wright, R.D. Nowak, and M.A.T. Figueiredo,
\newblock ``Sparse reconstruction by separable approximation,''
\newblock {\em IEEE Trans. on Signal Processing}, vol. 57, pp.
  2479--2493, 2009.

\bibitem{boyd2011distributed}
S.~Boyd, N.~Parikh, E.~Chu, B.~Peleato, and J.~Eckstein,
\newblock ``Distributed optimization and statistical learning via the
  alternating direction method of multipliers,''
\newblock {\em Foundations and Trends in Machine Learning},
  vol. 3, pp. 1--122, 2011.

\bibitem{goldstein2009split}
T.~Goldstein and S.~Osher,
\newblock ``The split {Bregman} method for l1-regularized problems,''
\newblock {\em SIAM Jour. on Imaging Sciences}, vol. 2, pp. 323--343, 2009.

\bibitem{chambolle2011first}
A.~Chambolle and T.~Pock,
\newblock ``A first-order primal-dual algorithm for convex problems with
  applications to imaging,''
\newblock {\em Jour. of Math. Imaging and Vision}, vol. 40, pp.
  120--145, 2011.

\bibitem{bauschke2011convex} 
H.H. Bauschke and P.L. Combettes,
\newblock {\em Convex analysis and monotone operator theory in Hilbert spaces},
\newblock Springer, 2011.

\bibitem{barzilai1988two}
J.~Barzilai and J.M. Borwein,
\newblock ``Two-point step size gradient methods,''
\newblock {\em IMA Jour. of Numerical Analysis}, vol. 8, pp. 141--148, 1988.

\end{thebibliography}

\end{document}